\documentclass[journal]{IEEEtranTIE}

\usepackage{graphicx}
\usepackage{cite}
\usepackage{picinpar}
\usepackage{amsmath}
\usepackage{url}
\usepackage{flushend}
\usepackage[latin1]{inputenc}
\usepackage{colortbl}
\usepackage{soul}
\usepackage{multirow}
\usepackage{pifont}
\usepackage{color}
\usepackage{alltt}
\usepackage[hidelinks]{hyperref}
\usepackage{enumerate}
\usepackage{siunitx}
\usepackage{breakurl}
\usepackage{epstopdf}
\usepackage{pbox}
\usepackage{amssymb}

\begin{document}
	\title{	Practical Distributed Control for Cooperative Multicopters in Structured Free Flight Concepts }
	
	\author{
		\vskip 1em
		Rao Fu,
		Quan Quan, \emph{Member IEEE},
		Mengxin Li,
		and Kai-Yuan Cai
		
		\thanks{		
			R. Fu, Q. Quan,  M. Li and K-Y. Cai are with the School of Automation Science and Electrical Engineering, Beihang University, Beijing 100191, China (e-mail:  buaafurao@buaa.edu.cn; qq\_buaa@buaa.edu.cn; lmxin@buaa.edu.cn; kycai@buaa.edu.cn).
		}
	}
	
	\maketitle
	
	\begin{abstract}
		Unmanned Aerial Vehicles (UAVs) are now becoming increasingly accessible to amateur and commercial users alike. Several types of airspace structures are proposed in recent research, which include several structured free flight concepts. In this paper, for simplicity, distributed coordinating the motions of multicopters in structured airspace concepts is focused. This is formulated as a free flight problem, which includes convergence to destination lines and inter-agent collision avoidance. The destination line of each multicopter is known a priori. Further, Lyapunov-like functions are designed elaborately, and formal analysis and proofs of the proposed distributed control are made to show that the free flight control problem can be solved. What is more, by the proposed controller, a multicopter can keep away from another as soon as possible, once it enters into the safety area of another one. Simulations and experiments are given to show the effectiveness of the proposed method.  
	\end{abstract}
	
	
	\markboth{IEEE TRANSACTIONS ON INDUSTRIAL ELECTRONICS}%
	{}
	
	\definecolor{limegreen}{rgb}{0.2, 0.8, 0.2}
	\definecolor{forestgreen}{rgb}{0.13, 0.55, 0.13}
	\definecolor{greenhtml}{rgb}{0.0, 0.5, 0.0}

\section{Introduction}

\begin{figure}
\begin{centering}
\includegraphics[scale=0.3]{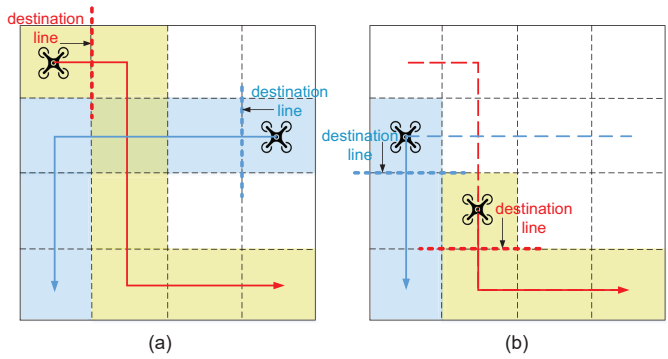} 
\par\end{centering}
\caption{Flying from a region to another under the zones-based airspace concept,
each UAV should switch its destination line corresponding to the next
region to complete its route.}
\label{concept} 
\end{figure}

\IEEEPARstart{A}IRSPACE is utilized today by far lesser aircraft
than it can accommodate, especially low-altitude airspace. There are
more and more applications for Unmanned Aerial Vehicles (UAVs) in
low-altitude airspace, ranging from the on-demand package delivery
to traffic and wildlife surveillance, inspection of infrastructure,
search and rescue, agriculture, and cinematography. Moreover, since
UAVs are usually small owing to portability requirements, it is often
necessary to deploy a team of UAVs to accomplish specific missions.
All these applications share a common need for both navigation and
airspace management. One good starting point is NASA's Unmanned Aerial
System Traffic Management (UTM) project, which organized a symposium
to begin preparations of a solution for low-altitude traffic management
to be proposed to the Federal Aeronautics Administration (FAA). What
is more, the design of Low-Altitude Air city Transport (LAAT) systems
has attracted more and more research \cite{IoD(2016)},\cite{Devasia(2016)}.
Several centralized and decentralized control approaches are proposed
for LAAT systems. A conclusion is that centralized architecture is
suitable for route planning and traffic flow control but lacks scalability
for conflict detection and collision avoidance \cite{Xue2020}; in
other words, the computational complexity is higher to solve a large
amount of conflicts among UAVs by centralized programming-based methods
\cite{Xue2019}. To address such a problem, free flight is a developing
air traffic control method that uses decentralized control \cite{Freeflight2000}.
Parts of airspace are reserved dynamically and automatically in a
distributed way using computer communication for separation assurance
among aircraft. This new system may be implemented into the U.S. air
traffic control system in the next decade. Airspace may be allocated
temporarily by UTM system for a particular task within a given time
interval. In this airspace, these aircraft have to be managed to complete
their tasks, i.e., arrive at the specific region while avoiding collisions.
Moreover, different airspace structures are investigated in recent
research. In the Metropolis project, layers-, zones-, and tubes-based
airspace concepts are investigated experimentally to benefit the airspace
capacity \cite{Hoekstra2018}. In the AIRBUS's Skyways project, the
tubes-based airspace concepts are focused on. The regions called `virtual
tubes' are designed to enable Vertical TakeOff and Landing (VTOL)
UAVs flights over the cities \cite{Airbus}. Another airspace concept
similar to the road network called `sky highway' is proposed in \cite{Skyhighway2021},
where aircraft are only allowed inside the following three: \emph{airways},
\emph{intersections}, and \emph{nodes}. More specifically, \emph{airways}
play a similar role to roads or virtual tubes, \emph{intersections}
are formed by at least two airways, and\emph{ nodes} are the points
of interest reachable through an alternating sequence of airways and
intersections. It is worth pointing out that the temporary target
of each UAV is always \emph{a chain of lines} or\emph{ planes }rather
than \emph{a chain of points} corresponding to the boundary of regions,
which is in contrast to the unstructured airspace concept. For example,
under the sky highway structure, the task of each UAV is to pass the
finish line of the airway at which it is located \cite{Skyhighway2021},\cite{Quan(2021)}.
Similarly, under the zones airspace concept \cite{Hoekstra2018},
the task of a UAV is from its origin to another region while avoiding
collision with other UAVs. For each UAV, a feasible path will be given
a priori as a chain of regions by the centralized path planning algorithms
(e.g., A-star or Dijkstra algorithm). The UAV will choose the temporary
target as the boundary line or plane from the current to the next
region, as shown in Fig. \ref{concept}(a) and (b). 

In this paper, distributed coordinating the motions of multicopters
in low-altitude structured airspace is focused on. Within the VTOL
ability, an important ability that might be mandated by authorities
in high traffic areas such as lower altitude in the urban airspace
\cite{IoD(2016)}, multicopters are highly versatile and can perform
tasks in an environment with very confined airspace available to them.
The main problem here, called the \emph{free flight control problem},
is to coordinate the motions of distributed multiple multicopters
include convergence to destinations (a plane or a line) and inter-agent
collision avoidance, which is very common in practice. For example,
the free flight area can be farmland or an area for package delivery.
The scenario mentioned above is also applicable to mobile multi-robot
systems or swarm robots. Such coordination problems of multiple agents
have been addressed partly using different approaches, various stability
criteria, and numerous control techniques \cite{Qin2017},\cite{Ren(2011)},\cite{Shi2017},\cite{Hoy(2015)}
(e.g., formation control methods \cite{Oh(2015)},\cite{Do(2006)},\cite{Rezaee2013},\cite{Zhao(2018)},
Lyapunov-like function methods \cite{Dimarogonas(2006)},\cite{Panagou(2016)},\cite{Hernandez(2011)},\cite{Wang(2017)},\cite{Quan(2017)},
optimal control methods \cite{Xue2019},\cite{Yu2020}). It is worth
pointing out that these approaches have their own strengths and weaknesses.
For example, the formation control methods perform well in scenarios
where multiple UAVs have the same task but have limitations in LAAT
systems because of their dependence on communication stability and
connectivity among multicopters, which is in contradiction with each
UAV performing its own task. The optimal control method trades for
optimal objectives (the cost of time, distance, or energy) at the
expense of time complexity using Linear Programming (LP) or Mixed-Integer
Linear Programming (MILP) algorithms, which is more suitable for centralized
control but lacks scalability for increasing UAVs.

Based on the reasons above, the proposed problem is mainly solved
using Lyapunov-like function methods in this paper because of its
ease of use and low time complexity. The control laws use the negative
gradient of mixing of attractive Lyapunov functions and barrier functions
to produce vector fields that ensure convergence and conflict avoidance,
respectively. It is similar to the Artificial Potential Field (APF)
based methods. However, for such type method, the deadlock and livelock
will exist, namely undesired equilibria appear. \textit{\emph{One
conclusion is stated in \cite{Rimon1992} that true global convergence
is not achieveable under APF based methods, i.e., there must exist
additional undesired }}equilibria; further,\textit{\emph{ Rimon-Koditschek
sense is proposed as a design principle for Lyapunov-like functions
to avoid collision for single agent with obstacles, which implies
that all undesired local minima disappear.}}\textit{ }\textit{\emph{This
implies that global convergence is achievable with probability 1,
namely deadlock avoidance is ensured. However, the limitation is that
livelock may happen under cooperative multi-agent cases. }}\textit{This
is the first problem for distributed coordination with only partial
information.}

Besides this problem, \textit{the second problem about conflict }\emph{resolution}\textit{
will also be encountered in practice.} The conflict between two agents
is often defined in control strategies that their distance is less
than a safety distance. In most literature, under the condition that
the initial distance among agents is more than the safety distance,
conflict avoidance among agents is proved formally rather than conflict
resolution \cite{Panagou(2016)},\cite{Quan(2017)}. However, a conflict
will happen in practice because of uncertainties such as estimated
noise, communication delay, and control delay. Due to the limitation
of the designed barrier function's domain, these strategies cannot
handle the $\left\Vert \mathbf{p}_{i}-\mathbf{p}_{j}\right\Vert <R$
situations ($\mathbf{p}_{i}$, $\mathbf{p}_{j}$ are two multicopters'
positions, and $R>0$ is the defined safety distance). This is a big
difference from some indoor robots with a highly accurate position
estimation and control. For such a similar problem, in \cite{Robustness},
a barrier function is proposed for controlling a nonlinear system
to operate within the safe set, but also outside the safe set with
some robustness margin. In \cite{noncooperative2021}, a preliminary
designed controller for multicopter is investigated to avoid a singer
non-cooperative moving obstacle, but the conclusion also has limitations
to extend to the case of multiple moving obstacles.

Motivated by the two problems, a distributed controller is proposed
to solve the free flight control problem for multiple cooperatives
multicopters in low-altitude structured airspace. The contributions
lie on the following properties of the proposed method.
\begin{itemize}
\item \textit{Neighboring information} \emph{used} \emph{without ID required}.
In practice, active detection devices such as cameras can only detect
neighboring multicopters' position and velocity but no IDs, because
these multicopters may look similar. Under this case, the proposed
controller can still work without considering the fixed topology,
which is quite different from the formation control methods.
\item \emph{Practical model used}. A kinematic model with the given velocity
command as input is proposed for multicopters. Compared to the single
or double integrator, the maneuverability for each multicopter has
been taken into consideration in this model. This model is simple
and easy to obtain in practice. What is more, distributed control
is developed for various tasks based on commercial semi-autonomous
autopilots. 
\item \emph{Control saturation}. The maximum velocity command in the proposed
distributed controller is confined according to the requirement of
semi-autonomous autopilots. Moreover, the maximum speed for each multicopter
approaching its destination line is further saturated so that the
contribution to the velocity command will not be dominated by the
term of approaching to destination line in the case of a multicopter
is very close to another. This avoids a danger that multicopters start
to change the velocity to avoid conflict too late.
\item \emph{Conflict-free} \emph{under extreme situations}. Formal proofs
about conflict avoidance are given. Moreover, the designed controller
has a larger domain; even if a multicopter enters into the safety
area of another multicopter, it can keep away from the neighboring
multicopters rapidly.
\item \emph{Convergence}. Formal proofs about the convergence for multiple
multicopters to the desired destination lines without deadlock are
given.
\item \emph{Low time complexity}. The proposed control protocol is simple
and can be computed at high speed, which is more suitable for increasing
agents than other approaches.
\end{itemize}

\section{Problem Formulation}

In this section, a multicopter control model is introduced first,
including the position model, the filtered position model, and the
safety radius model. For simplicity, these models are considered under
the 2-dimensional case. Then, the free flight control problem is formulated.

\subsection{Multicopter Control Model}

\subsubsection{Position Model}

There are $M$ multicopters in local airspace at the same altitude
satisfying the following model \cite{Quan(2021)},\cite{noncooperative2021}
\begin{align}
\mathbf{\dot{p}}_{i} & =\mathbf{v}_{i}\nonumber \\
\mathbf{\dot{v}}_{i} & =-l_{i}\left(\mathbf{v}_{i}-\mathbf{v}_{\text{c},i}\right)\label{positionmodel_ab_con_i}
\end{align}
where $\mathbf{p}_{i}\in\mathbb{R}^{2}$ , $\mathbf{v}_{i}\in{{\mathbb{R}}^{2}}$,
$\mathbf{v}_{\text{c},i}\in{{\mathbb{R}}^{2}}$ and $l_{i}>0$ are
the position, velocity, velocity command and horizontal control gain
of the $i$th multicopter respectively, $i=1,2,\cdots,M.$ This model
can also be adopted when a VTOL UAV takes flight with the altitude
hold mode. Similarly, the destination of the $i$th multicopter is
a line called the destination line as shown in Figure \ref{concept}(a)
and (b), which is defined as
\[
\mathcal{L}_{i}=\left\{ \mathbf{x}\in\mathbb{R}^{2}\left|\left(\mathbf{x}-\mathbf{p}_{\text{l,}i}\right)^{\text{\text{T}}}\mathbf{n}_{i}=0\right.\right\} 
\]
where $\mathbf{p}_{\text{l,}i}\in\mathbb{R}^{2}$ is a point located
at $\mathcal{L}_{i}$, and $\mathbf{n}_{i}$ denotes the unit normal
vector of $\mathcal{L}_{i}$. The control gain $l_{i}$ indicates
the maneuverability of the $i$th multicopter, which depends on the
semi-autonomous autopilot and can be obtained through flight experiments.
From the model (\ref{positionmodel_ab_con_i}), $\lim_{t\rightarrow\infty}\left\Vert \mathbf{v}_{i}\left(t\right)-\mathbf{v}_{\text{c},i}\right\Vert =0$
if $\mathbf{v}_{\text{c},i}$ is constant. Considering $v_{\text{m},i}>0$
is the maximum speed of the $i$th multicopter. The velocity command
$\mathbf{v}_{\text{c},i}$ for the $i$th multicopter is subject to
a saturation function defined as

\begin{equation}
\text{sat}\left(\mathbf{v},v_{\text{m},i}\right)=\kappa_{v_{\text{m},i}}\left(\mathbf{v}\right)\mathbf{v}\label{sat0}
\end{equation}
where $\mathbf{v}\triangleq\left[\begin{array}{cc}
v_{1} & v_{2}\end{array}\right]{}^{\text{T}}\in\mathbb{R}^{2}$, and

\begin{equation}
\kappa_{v_{\text{m},i}}\left(\mathbf{v}\right)\triangleq\left\{ \begin{array}{c}
1,\\
\frac{v_{\text{m},i}}{\left\Vert \mathbf{v}\right\Vert },
\end{array}\begin{array}{c}
\left\Vert \mathbf{v}\right\Vert \leq v_{\text{m},i}\\
\left\Vert \mathbf{v}\right\Vert >v_{\text{m},i}
\end{array}\right..\label{kwm}
\end{equation}
Without loss of generality, the Euclidean norm is used in the definition
of saturation function $\text{sat}\left(\mathbf{v},v_{\text{m},i}\right)$.
Note that $\text{sat}\left(\mathbf{v},v_{\text{m},i}\right)$ and
the vector $\mathbf{v}$ are parallel all the time so the multicopter
can keep the same flying direction under the case $\left\Vert \mathbf{v}\right\Vert >v_{\text{m},i}$
\cite[pp.260-261]{Quan(2017)}. It is obvious that $0<\kappa_{v_{\text{m},i}}\left(\mathbf{v}\right)\leq1$.
According to this, if and only if $\mathbf{v=0},$ then 
\begin{equation}
\mathbf{v}^{\text{T}}\text{sat}\left(\mathbf{v},v_{\text{m},i}\right)=0.\label{saturation1}
\end{equation}

\subsubsection{Filtered Position Model}

\vspace{-0.5cm}

\begin{figure}[h]
\begin{centering}
\includegraphics[scale=1.1]{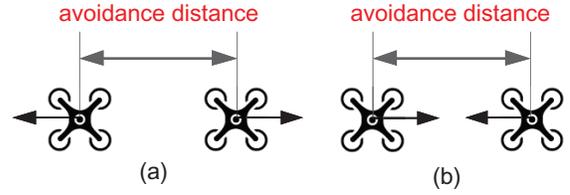} \vspace{-0.5em}
\par\end{centering}
\caption{Intuitive interpretation for the definition of filtered position.}
\label{Intuitive} 
\end{figure}

In this section, the motion of each multicopter is transformed into
a single integrator form to simplify the controller design and analysis.
As shown in Figure \ref{Intuitive}, although the position distances
are the same, namely a marginal avoidance distance, the case in Figure
\ref{Intuitive}(b) needs to carry out avoidance urgently by considering
the velocity. However, the case in Figure \ref{Intuitive}(a) does
not need to be considered to perform collision avoidance in fact.
With such an intuition, a filtered position is defined as follows:
\begin{equation}
\boldsymbol{\xi}_{i}\triangleq\mathbf{p}_{i}+\frac{1}{l_{i}}\mathbf{v}_{i}.\label{FilteredPosition}
\end{equation}
Then 
\begin{align}
\boldsymbol{\dot{\xi}}_{i} & =\mathbf{\dot{p}}_{i}+\frac{1}{l_{i}}\mathbf{\dot{v}}_{i}=\mathbf{v}_{\text{c},i}\label{filteredposdyn}
\end{align}
where $i=1,2,\cdots,M$. Define the position error and the filtered
position error between two multicopters as 
\begin{align*}
\mathbf{\tilde{p}}{}_{\text{m,}ij} & \triangleq\mathbf{p}_{i}-\mathbf{p}_{j}\\
\boldsymbol{\tilde{\xi}}{}_{\text{m,}ij} & \triangleq\boldsymbol{\xi}_{i}-\boldsymbol{\xi}{}_{j}.
\end{align*}
\emph{Proposition 1} \cite{Quan(2021)} indicates that the position
error is large enough as long as the filter position error is also
large enough, which is shown as follows:

\begin{equation}
\left\Vert \mathbf{\tilde{p}}{}_{\text{m,}ij}\left(t\right)\right\Vert \geq\left\Vert \boldsymbol{\tilde{\xi}}{}_{\text{m,}ij}\left(t\right)\right\Vert -\max_{i}\frac{v_{\text{m},i}}{l_{i}}.\label{filteredposition}
\end{equation}

\subsubsection{Safety Radius Model}

\begin{figure}[h]
\begin{centering}
\includegraphics[scale=0.9]{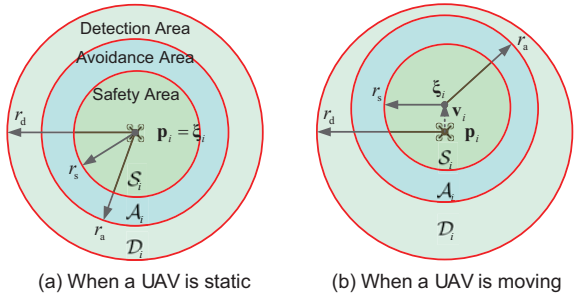} 
\par\end{centering}
\caption{Safety area, avoidance area and detection area of a multcopter \cite{Quan(2021)}.}
\label{Threeaera}
\end{figure}

Three types of areas used for control for the $i$th multicopter,
namely \emph{safety area} $\mathcal{S}_{i}$, \emph{avoidance area}
$\mathcal{A}_{i}$, and \emph{detection area} $\mathcal{D}_{i}$,
are defined, as shown in Figure \ref{Threeaera}. The\emph{ safety
area} $\mathcal{S}_{i}$ (to avoid a conflict) and \emph{avoidance
area} $\mathcal{A}_{i}$ (to start avoidance control) of the $i$th
multicopter are circles (spheres in 3-dimensional case) both centered
in its filtered position $\boldsymbol{\xi}_{i}$ with the \emph{safety}
\emph{radius} $r_{\text{s}}$ and the \emph{avoidance area }$r_{\text{a}}$,
respectively. In addition, the \emph{detection area }$\mathcal{D}_{i}$
only\emph{ }depends on the detection range of the used devices (by
cameras, radars, 4G/5G mobile, or Vehicle to Vehicle (V2V) communication),
which is centered in its position $\mathbf{p}_{i}$ with the \emph{detection}
\emph{radius} $r_{\text{d}}$. The specific design principles of the
safety radius are investigated in \cite{safetyradius2021}.

\textbf{Remark 1}. Intuitively, the basic design principle of safety
radius is guided in (\ref{filteredposition}). For two multicopters
satisfying the model (\ref{positionmodel_ab_con_i}), the error between
$\left\Vert \boldsymbol{\tilde{\xi}}{}_{\text{m,}ij}\left(t\right)\right\Vert $
and $\left\Vert \mathbf{\tilde{p}}{}_{\text{m,}ij}\left(t\right)\right\Vert $
has the upper bound $\max_{i}\frac{v_{\text{m},i}}{l_{i}}$. The condition
for obtaining this upper bound is that two multicopters move on the
same line and in opposite directions with maximum speed. To guarantee
safety in this extreme case, the safety radius should be at least
larger than the physical radius with $\max_{i}\frac{v_{\text{m},i}}{2l_{i}}$
for each multicopter. This indicates that the larger safety radius
should be designed for the multicopter with higher speed or lower
maneuverability if its physical radius is fixed.

\textbf{Remark 2}. It should be pointed out that the 2-dimensional
case is just for simplicity of description, while similar analysis
can also be extended to the 3-dimensional case. Specifically, the
model (\ref{positionmodel_ab_con_i}) can add the z-axis kinematic
function for each multicopter

\begin{align}
\dot{p}_{\text{z}} & =v_{\text{z}}\nonumber \\
\dot{v}_{\text{z}} & =-l_{\text{z}}\left(v_{\text{z}}-v_{\text{c,z}}\right)\label{positionmodel_ab_con_i-1}
\end{align}
where $p_{\text{z}},v_{\text{z}},v_{\text{c,z}}\in\mathbb{R}$ and
$l_{\text{z}}>0$ are the z-axis position, velocity, velocity command
and control gain of this multicopter, respectively. Note that the
z-axis control gain $l_{\text{z}}$ is different from the horizontal
control gain for multicopters in general. Further, the safety radius
model can also be extended to the 3-dimensional case, while the safety
area, avoidance area and detection area of a multicopter can be modeled
as a sphere, a cylinder, or an ellipsoid rather than a circle. As
for the stability analysis under the 3-dimensional case, similar proof
can be given.

\subsection{Problem Formulation}

The following assumptions are further needed. The position error and
the filtered position error between the $i$th multicopter and its
destination line $\mathcal{L}_{i}$ is defined as

\begin{align*}
\mathbf{\tilde{p}}{}_{\text{l,}i} & \triangleq\mathbf{A}_{\text{l,}i}\left(\mathbf{p}_{i}-\mathbf{p}_{\text{l,}i}\right)\\
\boldsymbol{\tilde{\xi}}{}_{\text{l,}i} & \triangleq\mathbf{A}_{\text{l,}i}\left(\boldsymbol{\xi}_{i}-\mathbf{p}_{\text{l,}i}\right)
\end{align*}
where $\mathbf{A}_{\text{l,}i}=\mathbf{n}_{i}\mathbf{n}_{i}^{\text{T}}$
is the projection operator \cite[p. 480]{Projection}. By (\ref{filteredposdyn}),
the derivative of the filtered errors above are 
\begin{align}
\boldsymbol{\dot{\tilde{\xi}}}{}_{\text{l,}i} & =\mathbf{A}_{\text{l,}i}\mathbf{v}_{\text{c},i}\label{wpmodel}\\
\boldsymbol{\dot{\tilde{\xi}}}{}_{\text{m,}ij} & =\mathbf{v}_{\text{c},i}-\mathbf{v}_{\text{c},j}\label{mmodel}
\end{align}
where $i\neq j$, $i,j=1,\cdots,M.$

\textbf{Assumption 1}. For each multicopter, the avoidance radius
satisfies $r_{\text{a}}>r_{\text{s}}$, and the detection radius satisfies
$r_{\text{d}}>r_{\text{s}}+r_{\text{a}}+2\underset{i}{\max}\frac{{v_{\text{m},i}}}{{l}_{i}}{.}$

\textbf{Assumption 2}. The multicopters' initial positions satisfy
\[
\left\Vert \boldsymbol{\tilde{\xi}}{}_{\text{m,}ij}\left(0\right)\right\Vert >2r_{\text{s}},i\neq j
\]
where $i,j=1,2,\cdots,M.$

\textbf{Assumption 3}. Mathematically, a multicopter arrives at its
destination line $\mathcal{L}_{i}$ if 
\begin{equation}
\left\Vert \mathbf{v}_{i}\right\Vert <\epsilon_{\text{a}}\text{ and }\left\Vert \mathbf{\tilde{p}}{}_{\text{l,}i}\right\Vert \leq\epsilon_{\text{d}}.\label{arrivial}
\end{equation}
where the sufficiently small $\epsilon_{\text{a}},\epsilon_{\text{d}}>0$
are given a priori. It implies that the multicopter arrives at the
next region. Further, the multicopter will switch its destination
line corresponding to its route, as shown in Figure \ref{concept}.

\textbf{Definition 1.} Let the set $\mathcal{N}_{\text{m},i}$ be
the collection of all mark numbers of other multicopters whose safety
aeras enter into the avoidance\emph{\ }area of the $i$th multicopter,
namely 
\[
\mathcal{N}_{\text{m},i}=\left\{ \left.j\right\vert \mathcal{S}_{j}\cap\mathcal{A}_{i}\neq\varnothing,j=1,\cdots,M,i\neq j\right\} .
\]
According to \textit{Assumption 1}, multicopters in $\mathcal{N}_{\text{m},i}$
can be detected by the \textit{i}th multicopter. For example, if the
safety areas of the $1$st, $2$nd multicopters enter into in the
avoidance area of the $3$rd multicopter, then $\mathcal{N}_{\text{m},3}=\left\{ 1,2\right\} $.

Based on \textit{Assumptions 1-}\emph{3}, for cooperative multicopters,
we have the \emph{free flight control problem} stated in the following.

\textbf{Objective}. Let $\mathbf{p}_{i}$ and the line $\mathcal{L}_{i}$
be be the position and the destination line of the $i$th multicopter,
respectively. Under \textit{Assumptions 1-}3, design the velocity
input $\mathbf{v}_{\text{c},i}$ for the $i$th multicopter with the
information of its neighboring set $\mathcal{N}_{\text{m},i}$ to
guarantee collision-avoidance and convergence to the destination line
$\mathcal{L}_{i}$, i.e., $\left\Vert \boldsymbol{\tilde{\xi}}{}_{\text{l,}i}\right\Vert $
converges to zero and $\left\Vert \boldsymbol{\tilde{\xi}}{}_{\text{m,}ij}\left(t\right)\right\Vert >2r_{\text{s}}$
holds for $t>0$, $i=1,\cdots,M$.

\begin{figure}
\begin{centering}
\includegraphics[scale=0.55]{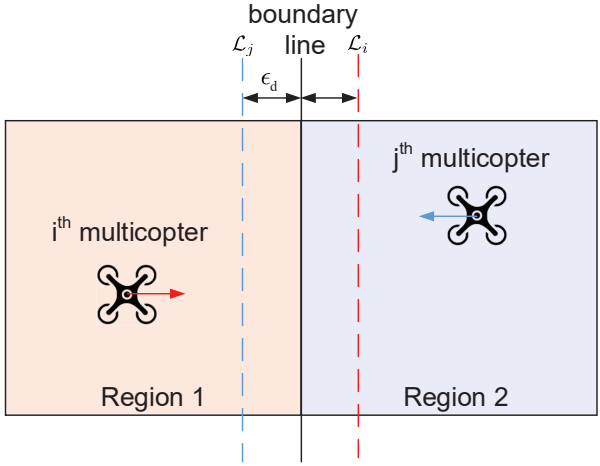} 
\par\end{centering}
\caption{The intuitive explanation for \textit{Assumption 3}. To achieve the
next region, the destination line for each multicopter can be chosen
parallel to the boundary line of two regions.}
\label{2regions} 
\end{figure}

\textbf{Remark 3}. According to \textit{Assumption 1},\textbf{ }for
the $i$th multicopter, any other multicopter entering into its avoidance
area can be detected by the $i$th multicopter and will not conflict
with the $i$th multicopter initially, $i=1,2,\cdots,M.$\textit{
Assumption 2} implies that any pair of two multicopters are not close
too much initially. \textit{Assumption 3} is also reasonable in practice
for air traffic, which is illustrated by the following example. Suppose
that the $i$th and $j$th multicopters are located at two adjacent
regions with the boundary line, while the task of each multicopter
is to arrive at another region, as shown in Figure \ref{2regions}.
To achieve this, the destination lines $\mathcal{L}_{i}$ and $\mathcal{L}_{j}$
can be chosen parallel to the boundary line of these two regions,
and the distance between $\mathcal{L}_{i}$, $\mathcal{L}_{j}$ and
the boundary line are both larger than $\epsilon_{\text{d}}$. Therefore,
$\left\Vert \mathbf{\tilde{p}}{}_{\text{l,}i}\right\Vert \leq\epsilon_{\text{d}}$
implies that the $i$th multicopter has arrived at the $j$th region
and the same for another multicopter. 

\section{Free Flight Control Problem for Multiple Cooperative Multicopters}

The idea of the proposed method is similar to that of the APF method.
In this method, the airspace is formulated as an APF. For a given
multicopter, only is the corresponding destination line assigned \textit{attractive
potential}, while other multicopters are assigned \textit{repulsive
potentials}. A multicopter in the field will be attracted to the destination
line, while being repelled by other multicopters.

\subsection{Preliminaries}

In the following, two designed smooth functions $\sigma\left(\cdot\right)$
and $s\left(\cdot\right)$ are used for the following Lyapunov-like
function design, which are defined as 
\begin{equation}
\sigma\left(x,d_{1},d_{2}\right)=\left\{ \begin{array}{c}
1\\
Ax^{3}+Bx^{2}+Cx+D\\
0
\end{array}\right.\begin{array}{c}
\text{if}\\
\text{if}\\
\text{if}
\end{array}\begin{array}{c}
x\leq d_{1}\\
d_{1}\leq x\leq d_{2}\\
d_{2}\leq x
\end{array}\label{zerofunction}
\end{equation}
with $A=-2\left/\left(d_{1}-d_{2}\right)^{3}\right.,$ $B=3\left(d_{1}+d_{2}\right)\left/\left(d_{1}-d_{2}\right)^{3}\right.,$
$C=-6d_{1}d_{2}\left/\left(d_{1}-d_{2}\right)^{3}\right.$ , $D=d_{2}^{2}\left(3d_{1}-d_{2}\right)\left/\left(d_{1}-d_{2}\right)^{3}\right.$
and \vspace{-0.5em}

\begin{equation}
s\left(x,\epsilon_{\text{s}}\right)=\left\{ \begin{array}{c}
x\\
\left(1-\epsilon_{\text{s}}\right)+\sqrt{\epsilon_{\text{s}}^{2}-\left(x-x_{2}\right)^{2}}\\
1
\end{array}\right.\begin{array}{c}
0\leq x\leq x_{1}\\
x_{1}\leq x\leq x_{2}\\
x_{2}\leq x
\end{array}\label{sat}
\end{equation}
with $x_{2}=1+\frac{1}{\tan67.5^{\circ}}\epsilon_{\text{s}}$ and
$x_{1}=x_{2}-\sin45^{\circ}\epsilon_{\text{s}}.$ The definition and
properties of these designed functions is analyzed in \cite{Quan(2021)}.
A new type of Lyapunov functions for vectors, called \emph{Line Integral
Lyapunov Function}, is designed as

\begin{equation}
V_{\text{l,i}}\left(\mathbf{y}\right)=\int_{C_{\mathbf{y}}}\text{sat}\left(\mathbf{x},a\right)^{\text{T}}\text{d}\mathbf{x}\label{Vli0}
\end{equation}
where $a>0,$ $\mathbf{x}\in\mathbf{\mathbb{R}}^{n},$ $C_{\mathbf{y}}$
is a line from $\mathbf{0}$ to $\mathbf{y}\in\mathbb{R}{}^{n}\mathbf{.}$
In the following lemma, we will show its properties.

\textbf{Lemma 1} \cite{Quan(2021)}. Suppose that the line integral
Lyapunov function $V_{\text{l,i}}$ is defined as (\ref{Vli0}). Then
(i) $V_{\text{l,i}}\left(\mathbf{y}\right)>0$ if $\left\Vert \mathbf{y}\right\Vert \neq0$;
(ii) if $\left\Vert \mathbf{y}\right\Vert \rightarrow\infty,$ then
$V_{\text{l,i}}\left(\mathbf{y}\right)\rightarrow\infty;$ (iii) if
$V_{\text{l,i}}\left(\mathbf{y}\right)$ is bounded, then $\left\Vert \mathbf{y}\right\Vert $
is bounded.

\subsection{Lyapunov-Like Function Design and Analysis}

Define a smooth curve $C_{\boldsymbol{\tilde{\xi}}{}_{\text{l,}i}}$
from $\mathbf{0}$ to $\boldsymbol{\tilde{\xi}}{}_{\text{l,}i}$.
Then, the line integral of $\text{sat}\left(\boldsymbol{\tilde{\xi}}{}_{\text{l,}i},v_{\text{m},i}\right)$
along $C_{\boldsymbol{\tilde{\xi}}{}_{\text{l,}i}}$ is 
\begin{equation}
V_{\text{l},i}\left(\boldsymbol{\tilde{\xi}}{}_{\text{l,}i}\right)=\int_{C_{\boldsymbol{\tilde{\xi}}{}_{\text{l,}i}}}\text{sat}\left(k_{1}\boldsymbol{\tilde{\xi}}{}_{\text{l,}i},v_{\text{m},i}\right)^{\text{T}}\text{d}\mathbf{x}\label{Vw}
\end{equation}
where $k_{1}>0,$ $i=1,2,\cdots,M$. Note that the reason of using
closed form of line integral in (\ref{Vw}) includes avoiding specifying
the norm in the saturation function (\ref{sat0}) and making the physical
meaning more intuitive. From the definition and \textit{Lemma 1},
$V_{\text{w},i}\geq0.$ Furthermore, define a barrier function as
\begin{equation}
V_{\text{m},ij}\left(\left\Vert \boldsymbol{\tilde{\xi}}{}_{\text{m,}ij}\right\Vert \right)=\frac{k_{2}\sigma_{\text{m}}\left(\left\Vert \boldsymbol{\tilde{\xi}}{}_{\text{m,}ij}\right\Vert \right)}{\left(1+\epsilon\right)\left\Vert \boldsymbol{\tilde{\xi}}{}_{\text{m,}ij}\right\Vert -2r_{\text{s}}s\left(\frac{\left\Vert \boldsymbol{\tilde{\xi}}{}_{\text{m,}ij}\right\Vert }{2r_{\text{s}}},\epsilon_{\text{s}}\right)}\label{Vmij}
\end{equation}
where $k_{2},\epsilon>0.$ Here $\sigma_{\text{m}}\left(x\right)\triangleq\sigma\left(x,2r_{\text{s}},r_{\text{a}}+r_{\text{s}}\right)$,
where $\sigma\left(\cdot\right)$ is defined in (\ref{zerofunction}).\ The
function $V_{\text{m},ij}$ has the following properties: (i) $\partial V_{\text{m},ij}\left/\partial\left\Vert \boldsymbol{\tilde{\xi}}{_{\text{m,}ij}}\right\Vert \right.\leq0$;
(ii) $\left\Vert \boldsymbol{\tilde{\xi}}{_{\text{m,}ij}}\right\Vert \geq r_{\text{a}}+r_{\text{s}}$
is a sufficient and necessary condition for $V_{\text{m},ij}=0$;
(iii) if $0<\left\Vert \boldsymbol{\tilde{\xi}}{_{\text{m,}ij}}\right\Vert <2r_{\text{s}},$
namely $\mathcal{S}_{j}\cap\mathcal{S}_{i}\neq\varnothing$ (they
may not collide in practice), then there exists a sufficiently small
$\epsilon_{\text{s}}>0$ such that 
\begin{equation}
V_{\text{m},ij}=\frac{k_{2}}{\epsilon\left\Vert \boldsymbol{\tilde{\xi}}{_{\text{m,}ij}}\right\Vert }=\frac{k_{2}}{2\epsilon r_{\text{s}}}.\label{Vmijd}
\end{equation}
The objective of the designed velocity command is to make $V_{\text{l},i}\left(\boldsymbol{\tilde{\xi}}{}_{\text{l,}i}\right)$
and $V_{\text{m},ij}\left(\left\Vert \boldsymbol{\tilde{\xi}}{}_{\text{m,}ij}\right\Vert \right)$
be zero or as small as possible. According to \textit{Lemma 1}\textbf{\ }and\textbf{\ }property
(ii), this implies $\left\Vert \boldsymbol{\tilde{\xi}}{}_{\text{l,}i}\right\Vert \rightarrow0$
and$\ \left\Vert \boldsymbol{\tilde{\xi}}{_{\text{m,}ij}}\right\Vert >r_{\text{a}}+r_{\text{s}}$,
namely the $i$th multicopter will arrive at the destination line
$\mathcal{L}_{i}$ and not conflict with the $j$th multicopter.

\subsection{Controller Design}

The velocity command is designed as 
\begin{equation}
\mathbf{v}_{\text{c},i}=-\text{sat}\left(\text{sat}\left(k_{1}\boldsymbol{\tilde{\xi}}{}_{\text{l,}i},v_{\text{m},i}\right)-\underset{j\in\mathcal{N}_{\text{m},i}}{\sum}b_{ij}\boldsymbol{\tilde{\xi}}{}_{\text{m,}ij},v_{\text{m},i}\right)\label{control_p3}
\end{equation}
where $i=1,2,\cdots,M.$ Here $b_{ij}=0,$ $i,j=1,\cdots,M,$ $i\neq j$
if (\ref{arrivial}) holds\footnote{It is used to represent that the $i$th multicopter quits the airspace.};
otherwise\footnote{$b_{ij}\geq0$ according to the property (i) of $V_{\text{m},ij}.$}
\begin{equation}
b_{ij}=-\frac{\partial V_{\text{m},ij}}{\partial\left\Vert \boldsymbol{\tilde{\xi}}{}_{\text{m,}ij}\right\Vert }\frac{1}{\left\Vert \boldsymbol{\tilde{\xi}}{}_{\text{m,}ij}\right\Vert }.\label{bij}
\end{equation}
In (\ref{control_p3}), the parameters $r_{\text{a}},r_{\text{s}}$
appear in $\sigma_{\text{m}}\left(x\right)$ included in (\ref{Vmij})
and further (\ref{bij}), where \emph{Assumption 1 }has to be satisfied.

\textbf{Remark 4}. The saturation the term $\text{sat}\left(k_{1}\boldsymbol{\tilde{\xi}}{}_{\text{l,}i},v_{\text{m},i}\right)$
in (\ref{control_p3})\ is very necessary in practice. Without the
saturation, the velocity command (\ref{control_p3})\ becomes 
\[
\mathbf{v}_{\text{c},i}=\mathbf{-}\text{sat}\left(k_{1}\boldsymbol{\tilde{\xi}}{}_{\text{l,}i}-\underset{j\in\mathcal{N}_{\text{m},i}}{\sum}b_{ij}\boldsymbol{\tilde{\xi}}{}_{\text{m,}ij},{v_{\text{m},i}}\right).
\]
In this case, if $\left\Vert \boldsymbol{\tilde{\xi}}{}_{\text{l,}i}\left(0\right)\right\Vert $
is very large, then the term $k_{1}\boldsymbol{\tilde{\xi}}{}_{\text{l,}i}$
will dominate until the multicopter is very close to another so that
$b_{ij}\boldsymbol{\tilde{\xi}}{_{\text{m,}ij}}$ will dominate. At
that time, the multicopter will start to change the velocity to avoid
the conflict. In practice, it may be too late by taking various uncertainties
into consideration. The use of the maximum speeds $v_{\text{m},i}$
in the term $\text{sat}\left(k_{1}\boldsymbol{\tilde{\xi}}{}_{\text{l,}i},v_{\text{m},i}\right)$
of the velocity command (\ref{control_p3}) will avoid such a danger.

\textbf{Remark 5}. It should be noted that, in most literature, if
their distance is less than a safety distance, then their control
schemes either do not work or even push the agent towards the center
of the safety area rather than leaving the safety area. Theses have
been explained in \textit{Introduction}. The proposed controller can
also handle the case such as $\left\Vert \boldsymbol{\tilde{\xi}}{_{\text{m,}ij_{i}}}\right\Vert <2r_{\text{s}}$,
which may still happen in practice due to unpredictable uncertainties.
However, this may not imply that the $i$th multicopter have collided
with the $j_{i}$th multicopter physically, because the redundancy
is always considered when we design the safety radius $r_{\text{s}}$,
i.e., $r_{\text{s}}$ is larger than the physical radius of the multicopter.
In this case, there exists a sufficiently small $\epsilon_{\text{s}}>0$
such that 
\[
b_{ij_{i}}=b_{j_{i}i}=\frac{k_{2}}{\epsilon}\frac{1}{\left\Vert \boldsymbol{\tilde{\xi}}{}_{\text{m,}ij_{i}}\right\Vert ^{3}}.
\]
Since $\epsilon$ is chosen to be sufficiently small, the terms $b_{ij_{i}}\boldsymbol{\tilde{\xi}}{}_{\text{m,}ij_{i}}$
and $b_{j_{i}i}\boldsymbol{\tilde{\xi}}{}_{\text{m,}j_{i}i}$ will
dominate so that the velocity commands $\mathbf{v}_{\text{c},i}$
and $\mathbf{v}_{\text{c},j_{i}}$ become 
\begin{align*}
\mathbf{v}_{\text{c},i} & =\text{sat}\left(\frac{k_{2}}{\epsilon}\frac{1}{\left\Vert \boldsymbol{\tilde{\xi}}{}_{\text{m,}ij_{i}}\right\Vert ^{2}}\frac{\boldsymbol{\tilde{\xi}}{}_{\text{m,}ij_{i}}}{\left\Vert \boldsymbol{\tilde{\xi}}{}_{\text{m,}ij_{i}}\right\Vert },v_{\text{m},i}\right)\\
\mathbf{v}_{\text{c},j_{i}} & =\text{sat}\left(\frac{k_{2}}{\epsilon}\frac{1}{\left\Vert \boldsymbol{\tilde{\xi}}{}_{\text{m,}j_{i}i}\right\Vert ^{2}}\frac{\boldsymbol{\tilde{\xi}}{}_{\text{m,}j_{i}i}}{\left\Vert \boldsymbol{\tilde{\xi}}{}_{\text{m,}j_{i}i}\right\Vert },v_{\text{m},j_{i}}\right).
\end{align*}
This implies that, by recalling (\ref{mmodel}), $\left\Vert \boldsymbol{\tilde{\xi}}{}_{\text{m,}ij_{i}}\right\Vert $
will be increased fast so that the $i$th multicopter and the $j_{i}$th
multicopter can keep away from each other immediately. This implies
that the proposed controller still works when \emph{Assumption 2}
is violated, which indicates the robustness of the proposed controller
and makes it more feasible in practice.

\subsection{Stability Analysis}

In order to investigate the convergence to the destination line and
the multicopter avoidance behaviour, a function is defined as follows
\begin{equation}
V_{1}=\underset{i=1}{\overset{M}{\sum}}\left(V_{\text{l},i}+\frac{{1}}{2}\underset{j=1,j\neq i}{\overset{M}{\sum}}V_{\text{m},ij}\right)\label{Lyapunovmultiply}
\end{equation}
where $V_{\text{l},i}$ is defined in (\ref{Vw}) and $V_{\text{m},ij}$
is defined in (\ref{Vmij}). According to Thomas' Calculus \cite[p. 911]{Thomas(2009)},
one has 
\[
V_{1}=\underset{i=1}{\overset{M}{\sum}}\left(\int\nolimits _{0}^{t}\text{sat}\left(k_{1}\boldsymbol{\tilde{\xi}}{}_{\text{l,}i},v_{\text{m},i}\right)^{\text{T}}\boldsymbol{\dot{\tilde{\xi}}}{}_{\text{l,}i}\text{d}\tau+\frac{1}{2}\underset{j=1,j\neq i}{\overset{M}{\sum}}V_{\text{m},ij}\right).
\]
The derivative of $V_{1}$ along the solution to {(\ref{wpmodel})
and (\ref{mmodel}) is 
\begin{align*}
\dot{V}_{1} & =\underset{i=1}{\overset{M}{\sum}}\left(\text{sat}\left(k_{1}\boldsymbol{\tilde{\xi}}{}_{\text{l,}i},v_{\text{m},i}\right)-\underset{j=1,j\neq i}{\overset{M}{\sum}}b_{ij}\boldsymbol{\tilde{\xi}}{}_{\text{m,}ij}\right)^{\text{T}}\mathbf{v}_{\text{c},i}
\end{align*}
where the property $b_{ij}=b_{ji}$ defined in (\ref{bij}) is used.
If $j\notin\mathcal{N}_{\text{m},i},$ one has $\mathcal{A}_{j}\cap\mathcal{S}_{i}=\varnothing.$
Then $b_{ij}=0$ according to the property (ii) of $V_{\text{m},ij}$.
Consequently, 
\[
\underset{j\in\mathcal{N}_{\text{m},i}}{\sum}b_{ij}\boldsymbol{\tilde{\xi}}{}_{\text{m,}ij}=\underset{j=1,j\neq i}{\overset{M}{\sum}}b_{ij}\boldsymbol{\tilde{\xi}}{}_{\text{m,}ij}.
\]
By using the velocity input (\ref{control_p3}), $\dot{V}_{1}$ becomes
\begin{align*}
\dot{V}_{1} & =-\underset{i=1}{\overset{M}{\sum}}\left(\text{sat}\left(k_{1}\boldsymbol{\tilde{\xi}}{}_{\text{l,}i},v_{\text{m},i}\right)-\underset{j\in\mathcal{N}_{\text{m},i}}{\sum}b_{ij}\boldsymbol{\tilde{\xi}}{}_{\text{m,}ij}\right)\\
 & \cdot\text{sat}\left(\text{sat}\left(k_{1}\boldsymbol{\tilde{\xi}}{}_{\text{l,}i},v_{\text{m},i}\right)-\underset{j\in\mathcal{N}_{\text{m},i}}{\sum}b_{ij}\boldsymbol{\tilde{\xi}}{}_{\text{m,}ij},v_{\text{m},i}\right)\\
 & \leq0.
\end{align*}

Further, the main result is stated.

\textbf{Theorem 1}. Under \textit{Assumptions 1-3}, suppose the velocity
command is designed as (\ref{control_p3}) for model (\ref{positionmodel_ab_con_i}).
Then there exist positive parameters $k_{1},k_{2},\epsilon,\epsilon_{\text{s}}>0$
in the proposed controller such that $\lim_{t\rightarrow\infty}\left\Vert \mathbf{\tilde{p}}_{\text{l,}i}\left(t\right)\right\Vert <\epsilon_{\text{d}}$
and $\left\Vert \boldsymbol{\tilde{\xi}}{}_{\text{m,}ij}\left(t\right)\right\Vert >2r_{\text{s}}$,
$t\in\left[0,\infty\right)$ for almost $\mathbf{\tilde{p}}_{\text{l,}i}(0)$,
$i\neq j,$ $i,j=1,2,\cdots,M$.

\textit{Proof}. Due to limited space, similar to \textit{Lemma 2}
\cite{Quan(2021)}, we can prove that these multicopters are able
to avoid conflict with each other, namely $\left\Vert \boldsymbol{\tilde{\xi}}{}_{\text{m,}ij}\left(t\right)\right\Vert >2r_{\text{s}}$,
$i\neq j,$ $i,j=1,2,\cdots,M.$ In the following, the reason why
each multicopter is able to arrive at the destination $\mathcal{L}_{i}$
is given. The \textit{invariant set theorem} \cite[p. 69]{Slotine(1991)}
is used to do the analysis\textit{.} 
\begin{itemize}
\item First, we will study the property of function $V_{1}$. Let $\Omega=\left\{ \left.\boldsymbol{\xi}_{1},\cdots\boldsymbol{\xi}_{M}\right\vert V_{1}\left(\boldsymbol{\xi}_{1},\cdots\boldsymbol{\xi}_{M}\right)\leq l\right\} ,$
$l>0.$ According to \textit{Lemma 2}, $V_{\text{m},ij}>0.$ Therefore,
$V_{1}\left(\boldsymbol{\xi}_{1},\cdots\boldsymbol{\xi}_{M}\right)\leq l$
implies $\underset{i=1}{\overset{M}{\sum}}V_{\text{l},i}\leq l.$
Furthermore, according to \textit{Lemma 1(iii)}, $\Omega$ is bounded.
When $\left\Vert \left[\begin{array}{ccc}
\boldsymbol{\xi}_{1} & \cdots & \boldsymbol{\xi}_{M}\end{array}\right]\right\Vert \rightarrow\infty,$ then $\underset{i=1}{\overset{M}{\sum}}V_{\text{l},i}\rightarrow\infty$
according to \textit{Lemma 1(ii)}, namely $V_{1}\rightarrow\infty.$ 
\item Secondly, we will find the largest invariant set. Then show all multicopters
can arrive at their corresponding destination lines. Now, recalling
the property (\ref{saturation1}), $\dot{V}_{1}$$=0$ if and only
if 
\begin{equation}
\text{sat}\left(k_{1}\boldsymbol{\tilde{\xi}}{}_{\text{l,}i},v_{\text{m},i}\right)-\underset{j\in\mathcal{N}_{\text{m},i}}{\sum}b_{ij}\boldsymbol{\tilde{\xi}}{}_{\text{m,}ij}=\mathbf{0}\label{equilibrium1_v}
\end{equation}
for $i=1,2,\cdots,M.$ Then $\mathbf{v}_{\text{c},i}=\mathbf{0}$
according to (\ref{control_p3}). Consequently, the equation (\ref{positionmodel_ab_con_i})
only holds if $\mathbf{v}_{i}=\mathbf{0}$ for $i=1,2,\cdots,M$.
Obviously, the equilibrium points are stable if $\left\Vert \boldsymbol{\tilde{\xi}}{}_{\text{l,}i}\right\Vert =0$
and $\left\Vert \boldsymbol{\tilde{\xi}}{}_{\text{m,}ij}\right\Vert >2r_{\text{a}}$,
$i,j=1,2,\cdots,M.$ The objective here is to prove that the other
equilibrium points $\left\Vert \mathbf{\tilde{p}}_{\text{l,}i}\right\Vert \geq\epsilon_{\text{d}}$
are unstable. According to (\ref{kwm}), define 
\begin{equation}
\kappa_{i}=\kappa_{v_{\text{m},i}}\left(k_{1}\boldsymbol{\tilde{\xi}}{}_{\text{l,}i}\right)=\left\{ \begin{array}{c}
1,\\
\frac{v_{\text{m},i}}{k_{1}\left\Vert \boldsymbol{\tilde{\xi}}_{\text{l,}i}\right\Vert },
\end{array}\begin{array}{c}
\left\Vert \boldsymbol{\tilde{\xi}}_{\text{l,}i}\right\Vert \leq\frac{v_{\text{m},i}}{k_{1}}\\
\left\Vert \boldsymbol{\tilde{\xi}}_{\text{l,}i}\right\Vert >\frac{v_{\text{m},i}}{k_{1}}
\end{array}\right..\label{ki}
\end{equation}
Note that the parameter $k_{1}$ can be sufficiently large such that
the relation $k_{1}>\frac{v_{\text{m},i}}{\epsilon_{\text{d}}}$ holds,
which implies that the input $k_{1}\mathbf{\tilde{p}}{}_{\text{l,}i}$
can keep saturated according to \emph{Assumption 3}, and only the
case $\left\Vert \boldsymbol{\tilde{\xi}}_{\text{l,}i}\right\Vert >\frac{v_{\text{m},i}}{k_{1}}$
in (\ref{ki}) should be considered. Then the equation (\ref{equilibrium1_v})
can be further written as 
\begin{align*}
 & \text{ sat}\left(k_{1}\boldsymbol{\tilde{\xi}}{}_{\text{l,}i},v_{\text{m},i}\right)-\underset{j\in\mathcal{N}_{\text{m},i}}{\sum}b_{ij}\boldsymbol{\tilde{\xi}}{}_{\text{m,}ij}\\
= & v_{\text{m},i}\frac{\boldsymbol{\tilde{\xi}}{}_{\text{l,}i}}{\left\Vert \boldsymbol{\tilde{\xi}}_{\text{l,}i}\right\Vert }-\underset{j\in\mathcal{N}_{\text{m},i}}{\sum}b_{ij}\boldsymbol{\tilde{\xi}}{}_{\text{m,}ij}=\mathbf{0}.
\end{align*}
Define 
\begin{equation}
\mathbf{v}_{\text{c},i}^{*}\triangleq v_{\text{m},i}\frac{\boldsymbol{\tilde{\xi}}{}_{\text{l,}i}}{\left\Vert \boldsymbol{\tilde{\xi}}_{\text{l,}i}\right\Vert }-\underset{j=1,j\neq i}{\overset{\bar{M}}{{\displaystyle \sum}}}b_{ij}\boldsymbol{\tilde{\xi}}{}_{\text{m,}ij}.\label{vci}
\end{equation}
For the ${\bar{M}}$ multicopters, substituting (\ref{control_p3})
into (\ref{filteredposdyn}) results in 
\begin{align}
\left[\begin{array}{c}
\boldsymbol{\dot{\xi}}_{1}\\
\mathbf{\vdots}\\
\boldsymbol{\dot{\xi}}_{\bar{M}}
\end{array}\right] & =\mathbf{f}\left(\boldsymbol{\xi}_{1},\cdots,\boldsymbol{\xi}_{\bar{M}}\right)=-\left[\begin{array}{c}
\kappa_{v_{\text{m},1}}^{\prime}\mathbf{v}_{\text{c},1}^{*}\\
\mathbf{\vdots}\\
\kappa_{v_{\text{m},\bar{M}}}^{\prime}\mathbf{v}_{\text{c},\bar{M}}^{*}
\end{array}\right]
\end{align}
where $\underset{j\in\mathcal{N}_{\text{m},i}}{\sum}b_{ij}\boldsymbol{\tilde{\xi}}_{\text{m},ij}=\underset{j=1,j\neq i}{\overset{\bar{M}}{{\displaystyle \sum}}}b_{ij}\boldsymbol{\tilde{\xi}}{}_{\text{m,}ij}$
is used and $\kappa_{v_{\text{m},i}}^{\prime}$ is defined as 
\begin{align}
\kappa_{v_{\text{m},i}}^{\prime} & =\left\{ \begin{array}{cc}
1, & \left\Vert \mathbf{v}_{\text{c},i}^{*}\right\Vert \leq v_{\text{m},i}\\
\frac{v_{\text{m},i}}{\left\Vert \mathbf{v}_{\text{c},i}^{*}\right\Vert }\text{,} & \left\Vert \mathbf{v}_{\text{c},i}^{*}\right\Vert >v_{\text{m},i}
\end{array}\right..\label{kvmi}
\end{align}
Note that $\left.\mathbf{v}_{\text{c},i}^{*}\right\vert _{\boldsymbol{\xi}_{i}=\mathbf{p}_{i}^{\ast}}=0$
holds at the equilibrium point $\boldsymbol{\xi}_{i}=\mathbf{p}_{i}^{\ast}$,
then $\left.\kappa_{v_{\text{m},i}}^{\prime}\right\vert _{\boldsymbol{\xi}_{i}=\mathbf{p}_{i}^{\ast}}=1$
holds. Then we can get the derivative of $\mathbf{f}\left(\boldsymbol{\xi}_{1},\boldsymbol{\xi}_{2},\cdots,\boldsymbol{\xi}_{\bar{M}}\right)$
with respect to $\left[\boldsymbol{\xi}_{1},\boldsymbol{\xi}_{2}\cdots,\boldsymbol{\xi}_{\bar{M}}\right]$
is 
\end{itemize}
\begin{eqnarray*}
\left.\frac{\partial\mathbf{f}\left(\boldsymbol{\xi}_{1},\boldsymbol{\xi}_{2},\cdots,\boldsymbol{\xi}_{\bar{M}}\right)}{\partial\left(\boldsymbol{\xi}_{1},\boldsymbol{\xi}_{2}\cdots,\boldsymbol{\xi}_{\bar{M}}\right)}\right\vert _{\boldsymbol{\xi}_{i}=\mathbf{p}_{i}^{\ast}} & = & \left.\left(\Lambda_{1}+\Lambda_{2}\right)\right\vert _{\boldsymbol{\xi}_{i}=\mathbf{p}_{i}^{\ast}}
\end{eqnarray*}

where the matrice $\mathbf{\Lambda}_{1},\Lambda_{2}$ are defined
in the following 
\begin{align*}
\mathbf{\Lambda}_{1} & =\left[\begin{array}{cccc}
b_{1} & -b_{12} & \cdots & -b_{1\bar{M}}\\
-b_{21} & b_{2} & \cdots & -b_{2\bar{M}}\\
\vdots & \vdots & \ddots & \cdots\\
-b_{\bar{M}1} & -b_{\bar{M}2} & \cdots & b_{\bar{M}}
\end{array}\right]\otimes\mathbf{I}_{2}\\
\mathbf{\Lambda}_{2} & =\left[\begin{array}{ccc}
\underset{j=2}{\overset{{\bar{M}}}{{\displaystyle \sum}}}\boldsymbol{\tilde{\xi}}{}_{\text{m,}1j}\frac{\partial b_{1j}}{\partial\boldsymbol{\xi}_{1}} & \cdots & -\boldsymbol{\tilde{\xi}}{}_{\text{m,}1\bar{M}}\frac{\partial b_{1\bar{M}}}{\partial\boldsymbol{\xi}_{\bar{M}}}\\
\vdots & \ddots & \vdots\\
-\frac{\partial b_{{\bar{M}}1}}{\partial\boldsymbol{\xi}_{1}}\boldsymbol{\tilde{\xi}}{}_{\text{m,}\bar{M}1}\frac{\partial b_{{\bar{M}}1}}{\partial\boldsymbol{\xi}_{1}} & \vdots & \underset{j=1}{\overset{\bar{M}-1}{{\displaystyle \sum}}}\boldsymbol{\tilde{\xi}}{}_{\text{m,}\bar{M}j}\frac{\partial b_{\bar{M}j}}{\partial\boldsymbol{\xi}_{\bar{M}}}
\end{array}\right].
\end{align*}
where $\otimes$ denotes Kronecker product, and the relationship
\[
\frac{\partial\left(\frac{\boldsymbol{\tilde{\xi}}{}_{\text{l,}i}}{\left\Vert \boldsymbol{\tilde{\xi}}{}_{\text{l,}i}\right\Vert }\right)}{\partial\boldsymbol{\xi}_{i}}=\frac{\partial\mathbf{n}_{i}}{\partial\boldsymbol{\xi}_{i}}=\mathbf{0}
\]
is utilized according to the definition of $\boldsymbol{\tilde{\xi}}{}_{\text{l,}i}$.
Note that the equilibrium point $\boldsymbol{\xi}_{i}=\mathbf{p}_{i}^{\ast}$
is unstable if and only if the matrix $\left.\Lambda\right\vert _{\boldsymbol{\xi}_{i}=\mathbf{p}_{i}^{\ast}}$
has at least one positive eigenvalue. By the definition of $b_{ij}$
in (\ref{bij}), the equation $\frac{\partial b_{ij}}{\partial\boldsymbol{\xi}_{j}}\boldsymbol{\tilde{\xi}}{_{\text{m,}ij}}=\frac{\partial b_{ji}}{\partial\boldsymbol{\xi}_{i}}\boldsymbol{\tilde{\xi}}{_{\text{m,}ji}}$
holds. Further, since $b_{ij}=b_{ji}$, the matrice $\Lambda_{1}$
and $\Lambda_{2}$ are both \emph{symmetric}. Note that $\Lambda_{1}$
has the form of Laplacian matrix of a directed graph since $b_{ij}>0$
holds, so it is a positive semidefinite matrix according to \emph{Lemma
1} \cite{Ren2005}. Further, define a column vector $\boldsymbol{\alpha}=\left[\begin{array}{ccc}
\left(\mathbf{R}\boldsymbol{\xi}_{1}\right)^{\text{T}} & \cdots & \left(\mathbf{R}\boldsymbol{\xi}_{\bar{M}}\right)^{\text{T}}\end{array}\right]^{\text{T}}$, where $\mathbf{R}=\left[\begin{array}{cc}
0 & 1\\
-1 & 0
\end{array}\right]$ is the rotation matrix. Note that $\boldsymbol{\alpha}^{\text{T}}\Lambda_{2}=\mathbf{0}$,
which implies that $\boldsymbol{\alpha}$ is the eigenvector of $\Lambda_{2}$
corresponding to zero eigenvalue. Then we have 
\begin{align*}
\boldsymbol{\alpha}^{\text{T}}\left.\left(\Lambda_{1}+\Lambda_{2}\right)\right\vert _{\boldsymbol{\xi}_{i}=\mathbf{p}_{i}^{\ast}}\boldsymbol{\alpha} & =\boldsymbol{\alpha}^{\text{T}}\left.\Lambda_{1}\right\vert _{\boldsymbol{\xi}_{i}=\mathbf{p}_{i}^{\ast}}\boldsymbol{\alpha}\\
 & =\underset{i=1}{\overset{\bar{M}}{{\displaystyle \sum}}}\underset{j=1,j\neq i}{\overset{\bar{M}}{{\displaystyle \sum}}}b_{ij}\left\Vert \boldsymbol{\tilde{\xi}}{}_{\text{m,}ij}\right\Vert ^{2}.
\end{align*}
This implies that one eigenvalue of $\left.\left(\Lambda_{1}+\Lambda_{2}\right)\right\vert _{\boldsymbol{\xi}_{i}=\mathbf{p}_{i}^{\ast}}$
at least has a positive real part. Therefore, the equilibrium point
$\boldsymbol{\xi}_{i}=\mathbf{p}_{i}^{\ast}$ is unstable, which is
in fact a saddle point (an intuitive explanation can be found in \cite[pp. 325-326]{Quan(2017)}),
$i=1,\cdots,\bar{M}$. For a saddle point, it is stable in a subspace
but unstable in the other space. The measure of the stable subspace
in the whole space equals $0$ or the stability probability is $0$.
Therefore, the equilibrium point $\mathbf{p}_{i}^{\ast}$ is unstable
with probability 1, i.e., any small deviation will drive the multicopter
away from $\mathbf{p}_{i}^{\ast}$. Therefore, $\lim_{t\rightarrow\infty}\left\Vert \mathbf{\tilde{p}}_{\text{l,}i}\left(t\right)\right\Vert <\epsilon_{\text{d}},$
$t\in\left[0,\infty\right)$. This complete the proof. $\square$

\textbf{Remark 6}. In \emph{Theorem 1}, \textit{\emph{the uncertainties
of systems are ignored, which implies the system is autonomous}}\textit{.
}Therefore, the condition of the \textit{invariant set theorem} is
satisfied to do the analysis\textit{\emph{. However, this does not
mean that our method is infeasible to the environment subject to uncertainties
such as noise, communication delay, packet loss, etc. A method is
proposed in \cite{safetyradius2021} with a principle that separates
the safety radius design and controller design. In other words, we
can design the controller under the ideal conditions and consider
all the uncertainties in the safety radius design process. The safety
radius should be larger than the physical radius of multicopters;
in other words, the margin of safety radius design should take uncertainties
into consideration. This also explains why the case }}$\left\Vert \boldsymbol{\tilde{\xi}}{_{\text{m,}ij_{i}}}\right\Vert <2r_{\text{s}}$
may happen in practice if the safety radius is designed inappropriately
or the uncertainties violate the assumption for the safety radius
design, as stated in \emph{Remark 5}.

\section{Simulation and Experiments}

Simulations and experiments are given in the following to show the
effectiveness of the proposed method, where a video about simulations
and experiments is available on https://youtu.be/NWysjgzBP6s.

\subsection{Numerical Simulation}

A scenario of a $250\text{m}\times250\text{m}$ square region is considered.
Each multicopter will enter the region from a random side of the square
with the safety radius $r_{\text{s}}=10$m, the avoidance radius $r_{\text{a}}=15$m,
the maximum speed $v_{\text{m}}=20\text{m/s}$, the control gain $l_{i}=5$,
and the destination line opposite to it entered. To show the effectiveness
of the proposed controller clearly, \emph{$M=420$} multicopters\emph{
}will enter the region with a dynamic inflow, which is shown in Figure
\ref{mindis}. A multicopter will be randomly placed on a boundary
line of the region it will enter, which implies that \emph{Assumption
2} may be violated and cause a conflict suddenly; however, the proposed
controller still works. This is to simulate the situation a multicopter
appears in another's safety area accidently. The snapshots of the
region are shown in Figure \ref{Position}, while the route of each
multicopter is plotted. By the proposed controller, each multicopter
will keep the safe distance larger than $2r_{\text{s}}=20$m with
other cooperative multicopters almost the whole process (except the
case that a confliction suddenly happens as the reason explained above).
Without loss of generality, the minimum distance $\underset{j\neq i,j=1,2,\cdots,M}{\min}\left\Vert \boldsymbol{\tilde{\xi}}{}_{\text{m,}ij}\right\Vert $
for $i=1,2,\cdots,40$ is shown in Figure \ref{mindis}(a). Note that
the minimum distance for a multicopter may be less than $2r_{\text{s}}=20$m
when it encounters another multicopter which just enters the region.
To indicate that each multicopter finally converges to its destination
line, the distance between the $i$th multicopter and its destination
line $\left\Vert \boldsymbol{\tilde{\xi}}{}_{\text{l,}i}\right\Vert $
for $i=1,2,\cdots,40$ is shown in Figure \ref{mindis}(b). The result
is consistent with the properties of the controller we proposed.

\begin{figure}
\begin{centering}
\includegraphics{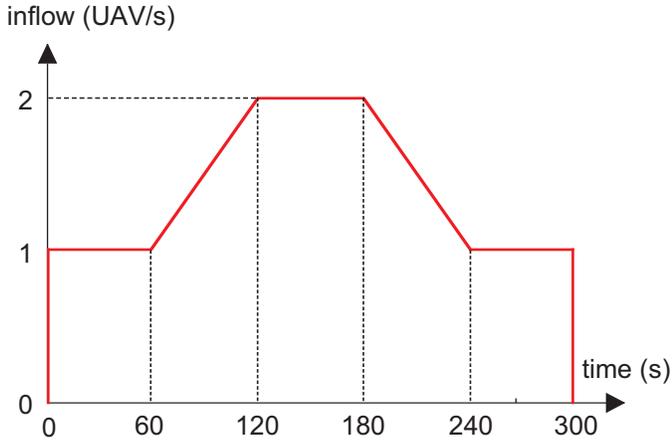} 
\par\end{centering}
\caption{The dynamic inflow process of the $250\text{m}\times250\text{m}$
square region.}
\label{inflow} 
\end{figure}

\begin{figure}
\begin{centering}
\includegraphics[scale=0.75]{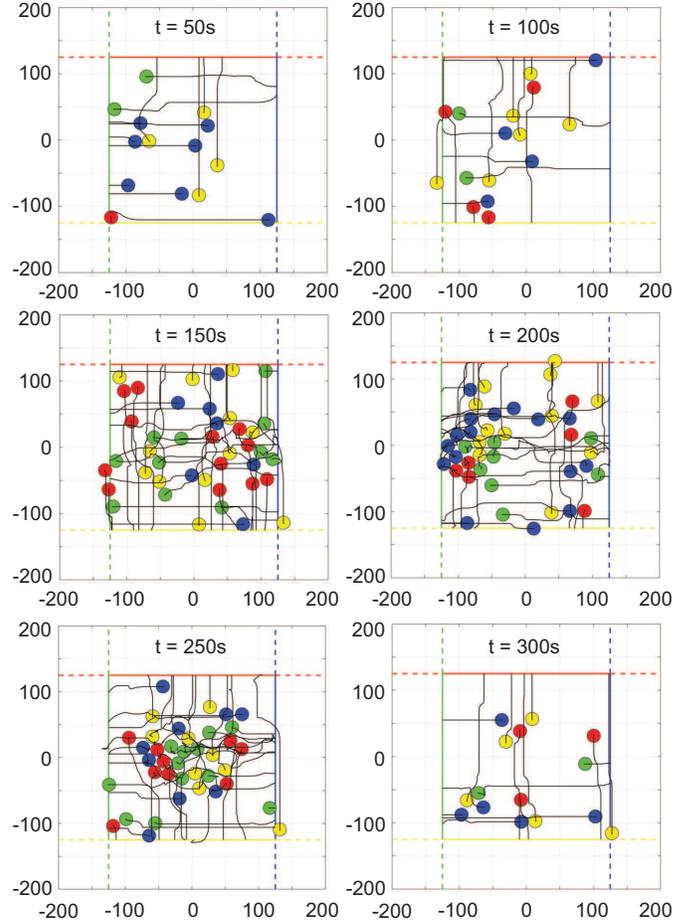} 
\par\end{centering}
\caption{Simulation snapshots of the 250m$\times$250m square region. The color
of each multicopter is consistent with its destination line.}
\label{Position} 
\end{figure}

\begin{figure}
\begin{centering}
\includegraphics[scale=0.55]{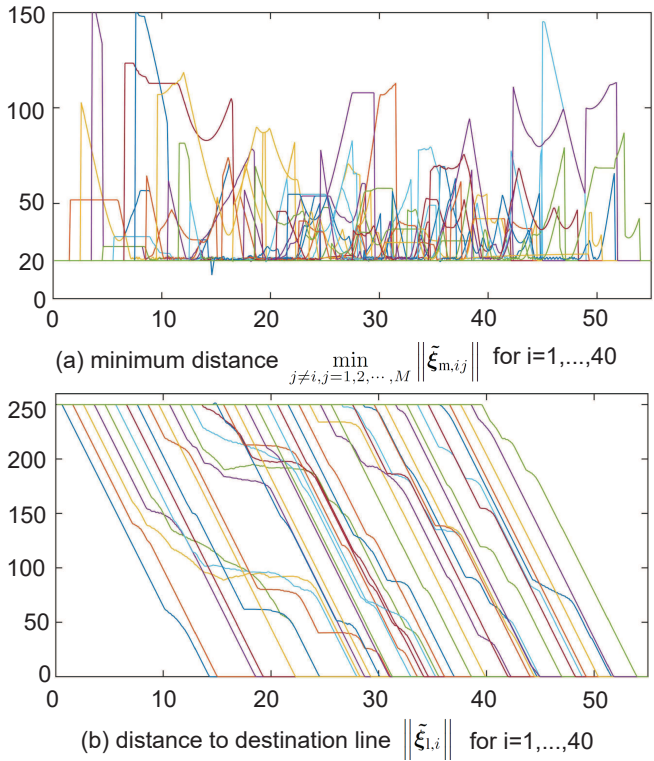}
\par\end{centering}
\caption{Minimum distance between the $i$th multicopter to others and distance
between the $i$th multicopter to its destination line for $i=1,2,\cdots,40$.}
\label{mindis} 
\end{figure}

\subsection{Experiments}

An indoor motion capture system called OptiTrack is installed in the
lab, from which we can get the ground truth of the position, velocity
and orientation of each multicopter. The laptop is running the proposed
controller on MATLAB 2020a. The laptop obtains the position and velocity
of each multicopter collected by optitrack through the local network,
and further controls the multicopters through the UDP protocol. Based
on the above conditions, a flight experiment is designed similarly
to the simulation scenario, which contains eight multicopters located
at four sides of a 2.5m$\times$2.5m square region initially. The
multicopters used for the experiment is Tello multicopters released
by DJI, where $r_{\text{s}}=0.2\text{m},$ $r_{\text{a}}=0.4\text{m},$
$v_{\text{m}}=0.15\text{m/s}$ are set. The destination line of each
multicopter is directly opposite to its origin. The positions and
the routes of multicopters during the whole flight experiment are
shown in Figure \ref{flightpos}. Finally, each multicopters can reach
its destination line at about $t=63$s, keeping a safe distance from
other multicopters without any conflict. 
\begin{figure}
\begin{centering}
\includegraphics[scale=0.65]{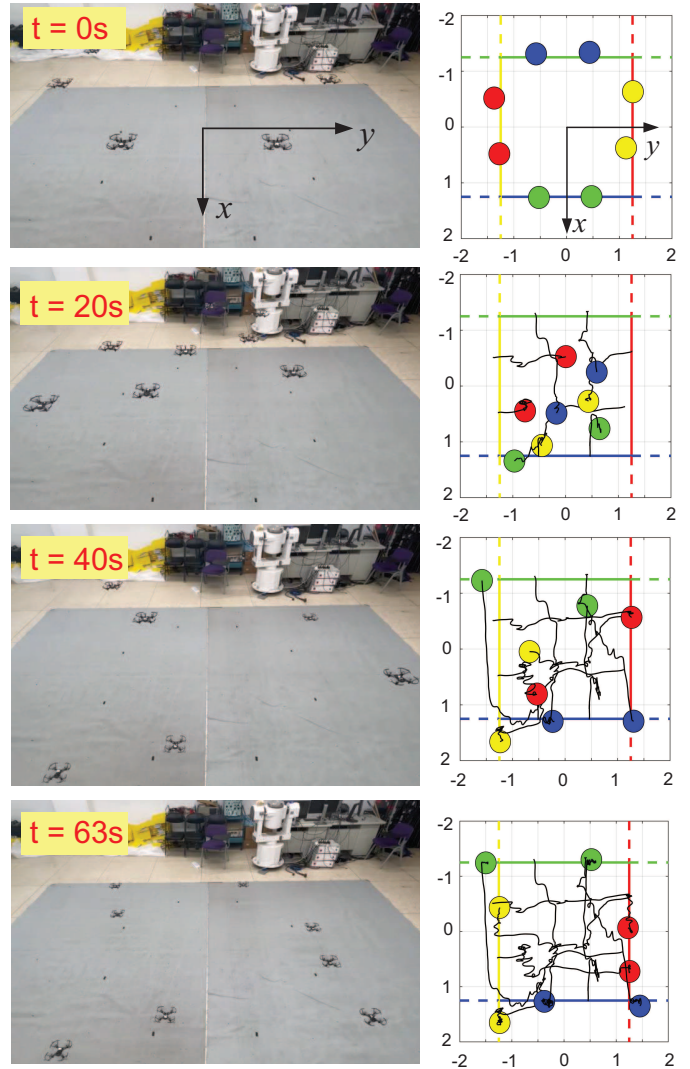} 
\par\end{centering}
\caption{Multicopters' positions at different time.}
\label{flightpos} 
\end{figure}

\subsection{Discussion}

The proposed control method can be easily extended to three-dimensional
situations. Different from most formation control methods, the proposed
method is more scalable and ensures that each agent completes its
own independent mission. The agent uses only the navigation information
of neighboring nodes to avoid potential collisions, so the topology
can be arbitrary rather than limited to a set. Compared to the optimization
based methods \cite{Ingersoll2016} and trajectory planning methods
\cite{Xian2020}, the proposed method is convenient to implement in
practical applications since it has low time complexity and avoids
the computation-consuming iterative optimization procedure. Moreover,
the designed controller has a larger domain than other Lyapunov-like
function methods \cite{Panagou(2016)},\cite{Quan(2017)}, which improves
safety under unpredictable uncertainties. Therefore, indoor robots
and air traffic are both included in potential applications.

\section{Conclusions}

The free flight control problem, which includes convergence to destination
line and inter-agent conflict avoidance with each multicopter, is
studied in this paper. Based on the velocity control model of multicopters
with control saturation, practical distributed control is proposed
for multiple multicopters to fly freely. Each multicopter has the
same and simple control protocol. Lyapunov-like functions are designed
with formal analysis and proofs showing that the free flight control
problem can be solved. Besides the functional requirement, the safety
requirement is also satisfied. By the proposed distributed control,
a multicopter can keep away from another as soon as possible, once
it enters into the safety area of another multicopter accidentally,
which is very necessary to guarantee safety. Simulations and experiments
are given to show the effectiveness of the proposed method from the
functional and safety requirements. 

\begin{IEEEbiography}[{\includegraphics[clip,width=1in,height=1.25in,keepaspectratio]{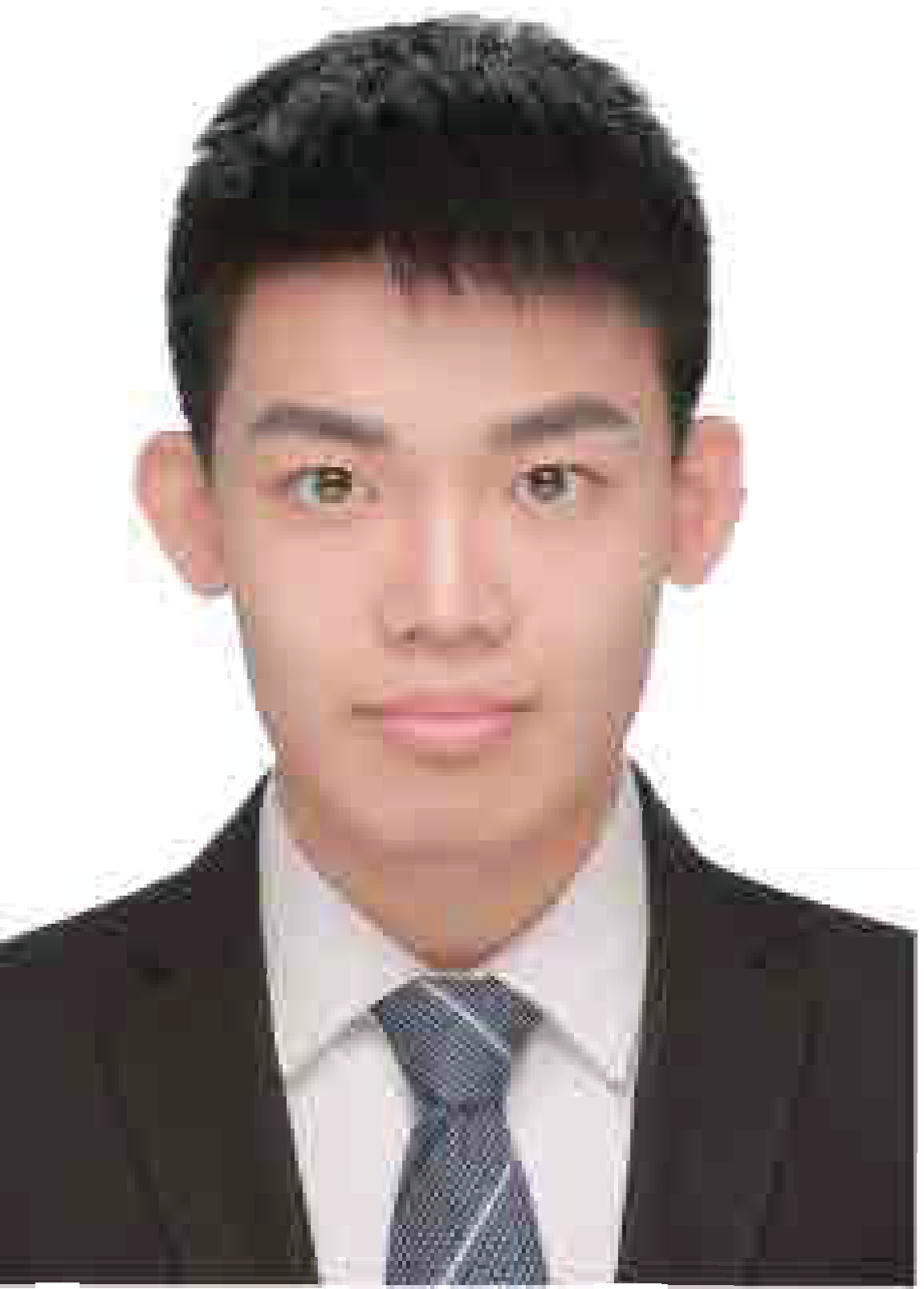}}]{Rao
Fu} received the B.S. degree in control science and engineering from
Beihang University, Beijing, China, in 2017. He is working toward
to the Ph.D. degree at the School of Automation Science and Electrical
Engineering, Beihang University (formerly Beijing University of Aeronautics
and Astronautics), Beijing, China. His main research interests include
UAV traffic control and swarm. \end{IEEEbiography}

\begin{IEEEbiography}[{\includegraphics[clip,width=1in,height=1.25in,keepaspectratio]{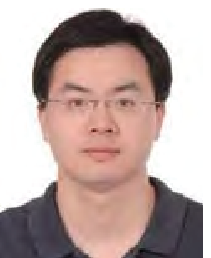}}]{Quan
Quan} received the B.S. and Ph.D. degrees in control science and
engineering from Beihang University, Beijing, China, in 2004 and 2010,
respectively. He has been an Associate Professor with Beihang University
since 2013, where he is currently with the School of Automation Science
and Electrical Engineering. His research interests include reliable
flight control, vision-based navigation, repetitive learning control,
and timedelay systems. \end{IEEEbiography} \begin{IEEEbiography}[{\includegraphics[clip,width=1in,height=1.25in,keepaspectratio]{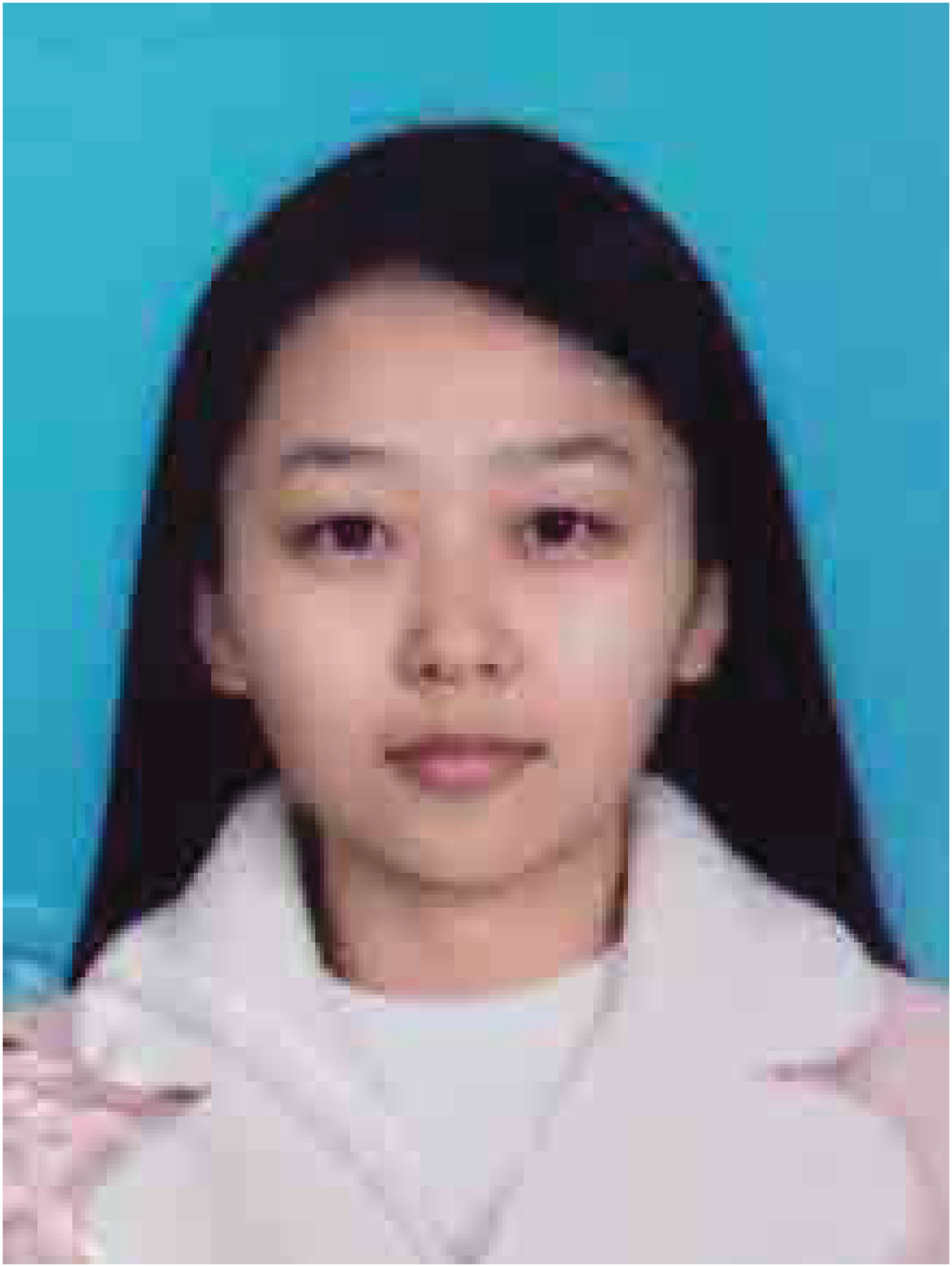}}]{Mengxin
Li} is working toward to the M.S. degree at the School of Automation
Science and Electrical Engineering, Beihang University (formerly Beijing
University of Aeronautics and Astronautics), Beijing, China. Her main
research interests include flight safety and control of multicopter.
\end{IEEEbiography} \begin{IEEEbiography}[{\includegraphics[clip,width=1in,height=1.25in,keepaspectratio]{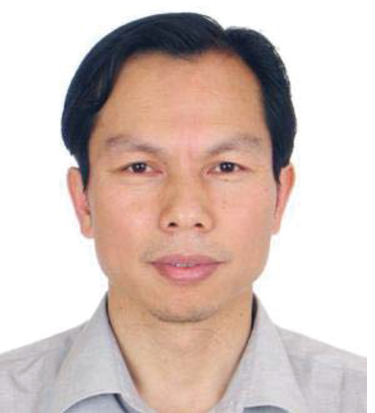}}]{Kai-Yuan
Cai} received the B.S., M.S., and Ph.D. degrees in control science
and engineering from Beihang University, Beijing, China, in 1984,
1987, and 1991, respectively. He has been a Full Professor at Beihang
University since 1995. He is a Cheung Kong Scholar (Chair Professor),
jointly appointed by the Ministry of Education of China and the Li
Ka Shing Foundation of Hong Kong in 1999. His main research interests
include software testing, software reliability, reliable flight control,
ADA (autonomous, dependable, and affordable) control, and software
cybernetics. \end{IEEEbiography} } 
\end{document}